# Technical Note: Feasibility of translating 3.0T-trained Deep-Learning Segmentation Models Out-of-the-Box on Low-Field MRI 0.55T Knee-MRI of Healthy Controls

**Bhattacharjee Rupsa,** Ph.D.[1], **Akkaya Zehra,** MD[1], **Luitjens Johanna**, MD[1], **Su Pan,** Ph.D.[2], **Yang Yang,** Ph.D.[1], **Pedoia Valentina,** Ph.D.[1,3], **Majumdar Sharmila,** Ph.D.[1]

1. Department of Radiology & Biomedical Imaging, University of California, San Francisco, USA
2. Siemens Medical Solutions USA Inc., Malvern, Pennsylvania, USA.
3. Image Analysis and Visualization Group, Altos Labs, San Francisco, USA

**Purpose:**

The significant contribution of Magnetic Resonance Imaging (MRI) in understanding disorders of the knee joint has been profound since the beginning of its clinical use. MR imaging of the knee has long been established in a plethora of conditions as well as research settings[1]. It is the most robust and non-invasive imaging modality enabling evaluation of all tissue components, including cartilage, ligaments and tendons, bone, and synovium. Osteoarthritis (OA) is a very costly and increasingly prevalent disease, which primarily affects the knee joint. Our understanding of this challenging disease has expanded with the use of MRI and the availability of reliable MRI biomarkers[2]. One of the most commonly used biomarkers is cartilage thickness, which is intricately related to bony anatomy as well as other factors. Although MRI is primarily used in the research field for knee OA, it is commonly employed in various clinical scenarios of the knee joint including post-traumatic or sports-related injuries. So far, most of the clinical utilities have been demonstrated on mid-field MRI systems, especially on 1.5T and 3.0T, considering the superior image quality, signal-to-noise ratio (SNR), and conventional availability catering to a larger population. In recent years, low-field (0.55T) MRI scanners have experienced a renaissance[3] with novel technical developments ensuring high-quality image acquisition with improved resolution, SNR, and accelerated scan time. Moreover, the 0.55T scanner offers additional advantages such as allowing safer image acquisition due to much lowered specific absorption rates (SAR) and providing higher flexibility to be installed at a broader geography due to lower weight and transportation requirements, with no quench pipe and minimal helium involved[4]. Potential uses of 0.55T scanners have been demonstrated for brain, pulmonary, cardiac, and musculoskeletal imaging domains [5][6][7]. Still in its early days, quantitative biomarker estimation and associated reliability have not been explored in-depth for the musculoskeletal system, particularly for knee OA. Designing newer DL algorithms for each specific field-strength segmentation problem would require a good amount of data acquisition and would eventually result in a long time of manual labor for radiologists in the meantime. There is a potential to utilize the existing DL algorithms out-of-the-box, which are extensively validated at higher-field strengths for specific tasks, for application in lower-field. In the current study, our purpose is to evaluate the feasibility of applying deep learning (DL) enabled algorithms to quantify bilateral knee biomarkers in healthy controls scanned at 0.55T and compared with 3.0T. The current study assesses the performance of standard in-practice bone, and cartilage segmentation algorithms at 0.55T, both qualitatively and quantitatively, in terms of comparing segmentation performance, areas of improvement, and compartment-wise cartilage thickness values between 0.55T vs. 3.0T.

**Methods:**

*Subjects:* In this ongoing prospective study, approved by the local Institutional Review Board (IRB), seven healthy control subjects (Age: 29.57 ± 5.25 years, BMI: 23.47 ± 2.74, 3 females), were recruited for simultaneous bilateral knee acquisitions on two MRI scanners. The subjects provided written informed consent before data collection. Inclusion criteria were a) being above 18 years of age, b) having no previous history of surgery on either knee, c) absence of clinically diagnosed knee OA, d) absence of a recent history of trauma in the past 3 months of recruitment, e) absence of any intraarticular injection in the past 6 months of recruitment, f) absence of sickle cell disease, hemoglobinopathy, inflammatory arthropathy, hematochromatosis, and contraindications to MRI. The subjects having fulfilled the required criteria, underwent MRI acquisitions between March 2023 and June 2023.

*MRI Acquisition:* All the participants underwent two consecutive MRI scans on a 3.0T GE-Signa Premier scanner (GE Healthcare, Waukesha, WI, USA) as well as on a 0.55T (MAGNETOM Free.Max, Siemens Healthineers, Erlangen, Germany). The subjects were given an hour of rest between undergoing both of these scans on the same day. The subjects were positioned supine, feet-first on both scanners. On 3.0T, simultaneous bilateral knee scans were performed with two 16-channel medium flex, receive-only coils (Neo-Coil, Pewaukee, WI, USA) wrapped around each knee. Simultaneous bilateral sagittal 3D proton-density fat-saturated fast-spin-echo (3D PDFS FSE i.e., Cube) were acquired with field of view (FOV) 160

× 160 mm², acquisition resolution 0.6 × 0.6 × 0.6 mm³, reconstruction resolution 0.3125 × 0.3125 × 0.6 mm³ and 548 slices for both knees, i.e., 224 slices per knee (TR/TE = 1202/26.62 ms). On 0.55T unilateral knee scans were performed for both left and right knees one at a time, wrapped in a Contour-S coil with 24 elements. Sagittal 3D PDFS, sampling perfection with application optimized contrast using different flip angle evolutions (Space) images were acquired with FOV 160 × 160 mm², acquisition resolution 0.6 × 0.6 × 0.6 mm³, reconstruction resolution 0.3125 × 0.3125 × 0.6 mm³ and 224 slices (TR/TE = 800/36 ms). In both the field strengths, the fat suppression was achieved via the spectral selective attenuation recovery fat suppression technique (SPAIR).

*Analysis:* For 3.0T images, the bilateral data were separated into left and right knees using an automated algorithm. Both the 3.0T Cube and 0.55T Space images underwent a modified 2D-convolutional neural networks (CNN) architecture[8] and a 3D V-Net architecture[9] respectively to segment the three major knee bones (Femur, Tibia, Patella) and their respective cartilage surfaces (Femoral, Tibial, Patellar cartilages). Both the DL architectures were previously trained and extensively validated on similar Cube image contrasts from another cohort scanned at 3.0T and were inferred on the 0.55T and 3.0T images in this study. Mean cartilage thickness values for three cartilage surfaces were automatically computed using an Euclidean distance transform[10]. All image analyses were performed using an in-house program developed in MATLAB (version R2021a, The MathWorks Inc., Natick, MA, USA) unless otherwise specified. The overall quality assessment was performed for fourteen knees from 7 participants acquired on the 0.55T as well as on the 3.0T by a musculoskeletal radiologist (ZA, with over nine years of experience). The bone and cartilage segmentation performances were assessed compartment-wise (for three major bones and cartilages) using a 5-point Likert scale via the following ranks: (1) unusable; (2) poor, with some mislabeling of bones or cartilage; (3) useable, with some major issues, but correct labeling of bone or cartilage; (4) good, with some minor but acceptable issues; (5) (near) human-like performance. Additionally, segmentation error scores were assigned to each of the segmentation as (0) no error; (1) missing (unsegmented) focal regions; (2) focal regions of extra segmentation; (3) general underestimation; (4) general overestimation; (5) combined errors (more than one type of error from 1-4). The overall image SNR was ranked as (0) sufficient or (1) insufficient. All statistical analyses have been performed using RStudio (version 12.0+353; https://www.r-project.org/). Likert scores[11] for segmentation quality, and cartilage thickness values per compartment are presented as mean ± standard deviation (SD). Three Shapiro-Wilk tests were performed to evaluate the normality of the difference of mean cartilage thickness computations (Femoral, Tibial, Patellar) from 0.55T and 3.0T. Three paired t-tests were conducted to determine whether, on average, there was a difference in cartilage thickness for Femoral, Tibial, and Patellar cartilages respectively, between 0.55T and 3.0T. The test statistic (t), degree of freedom (df), and 95% confidence intervals (CI) were reported. The significance threshold was set at alpha level $p \leq 0.05$.

**Results and Discussion:**

Without any sort of pre-training, as an initial inference run, both segmentation models were able to segment the three major bones and cartilage masks with reasonable ability, as demonstrated in Figure 1. Likert scores for the segmentations are summarized in Table 1. Tibial cartilage segmentation works better in 0.55T (score: good-to-near human-like) than in 3.0T (score: usable-to-good). For the Patellar cartilage, however, the segmentation quality at 3.0T (score: good-to-near human-like) was markedly better than 0.55T (score: poor-to-usable). Femoral cartilage segmentation performances were almost the same in both field strengths (score: good-to-near human-like). bone segmentations, in general, perform better in 3.0T (Tibia and Patella score: good-to-near human-like, Femur score: usable-to-good) than 0.55T (Patella score: good-to-near human-like, Tibia and Femur score: usable-to-good). The overall image SNR quality was rated sufficient for segmentation, for all the cases at 0.55T and 3.0T. Figure 2 summarizes the common issues with segmentation performance at 0.55T as well as 3.0T. The segmentation algorithm performance suffered mostly from missing a few spots on both 0.55T and 3.0T images. Likely with the addition of morpological postprocessing and refining, this limitation can be overcome.

The cartilage thickness values estimated for Femoral, Tibial, and Patellar cartilages for 0.55T and 3.0T are summarized in Table 2. From the Shapiro-Wilk test, the estimated p-values were 0.865 (Femoral), 0.5534 (Tibial), and 0.3799 (Patellar). The results imply the distributions of the differences (d) between mean Femoral, Tibial, and Patellar cartilage thickness calculated from 0.55T and 3.0T, are not significantly different from a normal distribution. In other words, we can assume the normality. However, the paired t-test results reject the null hypotheses and indicate that there were significant differences between cartilage thickness values at 0.55T vs. 3.0T: Femoral (t = -4.55, df = 13, 95% CI = [-0.412, -0.147], p<0.05), Tibial (t = -3.94, df = 13, 95% CI = [-0.449 -0.131], p<0.05), and Patellar (t = -4.55, df = 13, 95% CI = [-0.776 -

0.311], p<0.05), respectively. The mean differences between cartilage thickness quantifications at 3.0T vs. 0.55T were 0.279 (Femoral), 0.290 (Tibial), and 0.543 (Patellar) respectively. Comparing such lower mean difference estimates along with in-plane pixel resolutions (0.3125 × 0.3125 for Cube at 3.0T, and Space at 0.55T) implies that the majority of mismatch between the thickness values comes from an underestimation of a few couple pixels (less than 10) on 0.55T. The general missing spots/underestimations at 0.55T, also visualized in Figure 3, might quite reasonably be distributed across a 3D stack of image slices. The mean differences between cartilage thicknesses between 0.55T and 3.0T turn out significant also majorly due to the lower number of samples included in this study. With the inclusion of a higher sample size, such lower estimates of mean differences might likely wash out as non-significant.

In general, the cartilage segmentation algorithm outperformed the bone segmentation module at 0.55T, in the Femoral and Tibial regions, in terms of precision in detecting smaller cartilage regions .It could be hypothesized from the underestimation in cartilage thickness calculated from the Euclidean distance of the cartilage skeleton, that the misclassifications especially occur at the boundaries of cartilageneous and non-cartilagenous tissues. However, the cartilage segmentation modules suffered in the detection of the Patellar region. It is most likely due to insufficient training in low-SNR images seen in 0.55T. The 0.55T images suffer from inhomogeneous fat suppression at certain regions. The range of image intensities and tissue appearances are varied across field strengths as well. However, these experiments demonstrate a decent baseline of quantitative capabilities. There are further possibilities of improvement and modification of the 0.55T images to mimic the 3T images. This could be achieved either using more training data from 0.55T or by synthesizing similar-to-Cube (3T) images from the 0.55T image resolution and contrast utilizing super-resolution/generative network approaches[12]. Additionally, patellar cartilage could be better detected on an axial orientation. Even though the segmentation modules are trained to segment on sagittal acquisitions, an additional input of axially acquired images might benefit the segmentation.

**Conclusion:**

Initial results demonstrate a usable to good technical feasibility of translating existing quantitative deep-learning-based image segmentation techniques, trained on 3.0T, out of 0.55T for knee MRI, in a multi-vendor acquisition environment. Especially in terms of segmenting cartilage compartments, the models perform almost equivalent to 3.0T in terms of Likert ranking. The 0.55T low-field sustainable and easy-to-install MRI, as demonstrated, thus, can be utilized for evaluating knee cartilage thickness and bone segmentations aided by established DL algorithms trained at higher-field strengths out-of-the-box initially. This could be utilized at the far-spread point-of-care locations with a lack of radiologists available to manually segment low-field images, at least till a decent base of low-field data pool is collated. With further fine-tuning with manual labeling of low-field data or utilizing synthesized higher SNR images from low-field images, OA biomarker quantification performance is potentially guaranteed to be further improved.

**Tables and Figures:**

**Table 1:** Summarized Likert ratings for bone and cartilage segmentation quality assessment at 0.55T and 3.0T. 5-point Likert ranks defined are: (1) unusable; (2) poor, with some mislabeling of bones or cartilage; (3) useable, with some major issues, but correct labeling of bone or cartilage; (4) good, with some minor but acceptable issues; (5) (near) human-like.

| Region | Likert rating for bone segmentation (0- 5) mean ± standard deviation (SD) | | Likert rating for cartilage segmentation (0- 5) mean ± standard deviation (SD) | |
|---|---|---|---|---|
| | **0.55T** | **3.0T** | **0.55T** | **3.0T** |
| Femur bone / Femoral cartilage | 3.5 ± 0.85 | 3.64 ± 0.84 | 4.00 ± 0.55 | 4.07 ± 0.91 |
| Tibia bone / Tibial cartilage | 3.85 ± 0.77 | 4.14 ± 0.53 | 4.5 ± 0.65 | 3.78 ± 1.05 |
| Patella bone /Patellar cartilage | 4.14 ± 0.94 | 4.35 ± 0.49 | 2.57 ± 1.5 | 4.42 ± 0.51 |

**Table 2:** Cartilage Thickness Values computed Based on Automated Cartilage Segmentations (in millimeters) on Seven volunteers (fourteen knees)

| Compartment | Cartilage thickness values (mean ± SD) (mm) | |
|---|---|---|
| | **3.0T** | **0.55T** |
| Femoral | 3.628 ± 0.37 | 3.469 ± 0.28 |
| Tibial | 2.904 ± 0.65 | 2.786 ± 0.50 |
| Patellar | 3.023 ± 0.91 | 3.124 ± 0.63 |

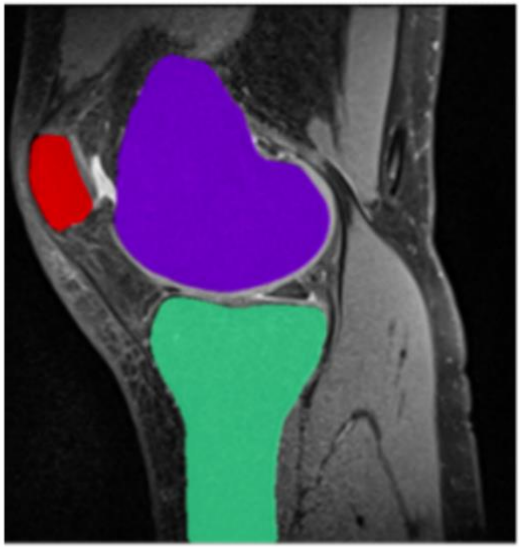 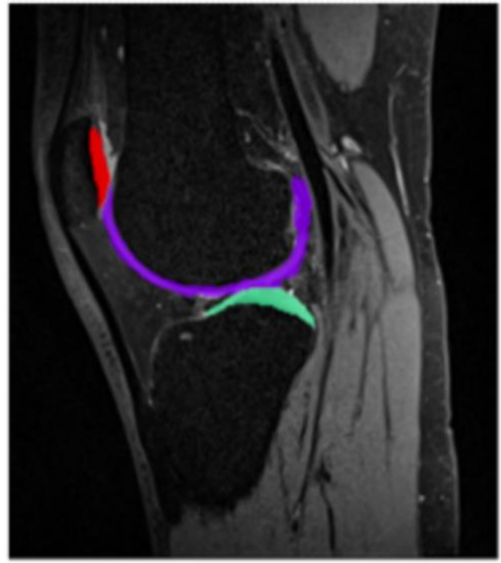
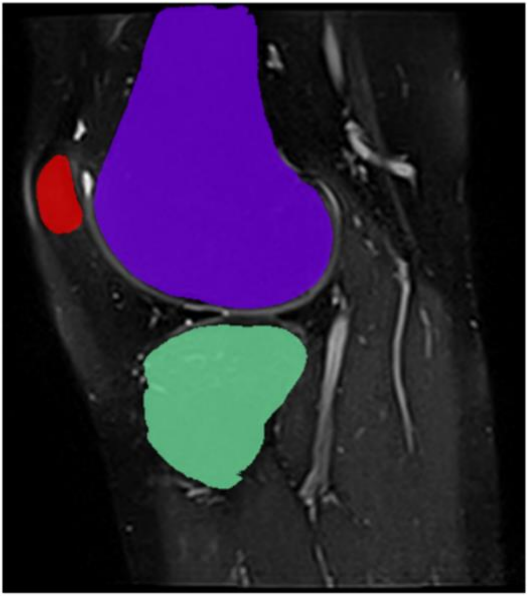 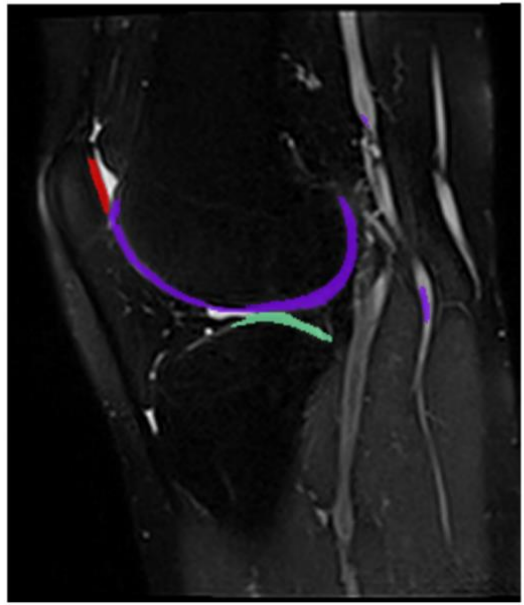

**Figure 1**: 3.0T: (A) Three major bone (Femur, Tibia, Patella) segmentations, (B) three major cartilage (Femoral, Tibial, Patellar) segmentations overlaid on 3D-PD Fat Suppressed Cube images. 0.55T: (C) Three major bone and (B) three major cartilage segmentations overlaid on 3D-PD Fat-Suppressed Space images.

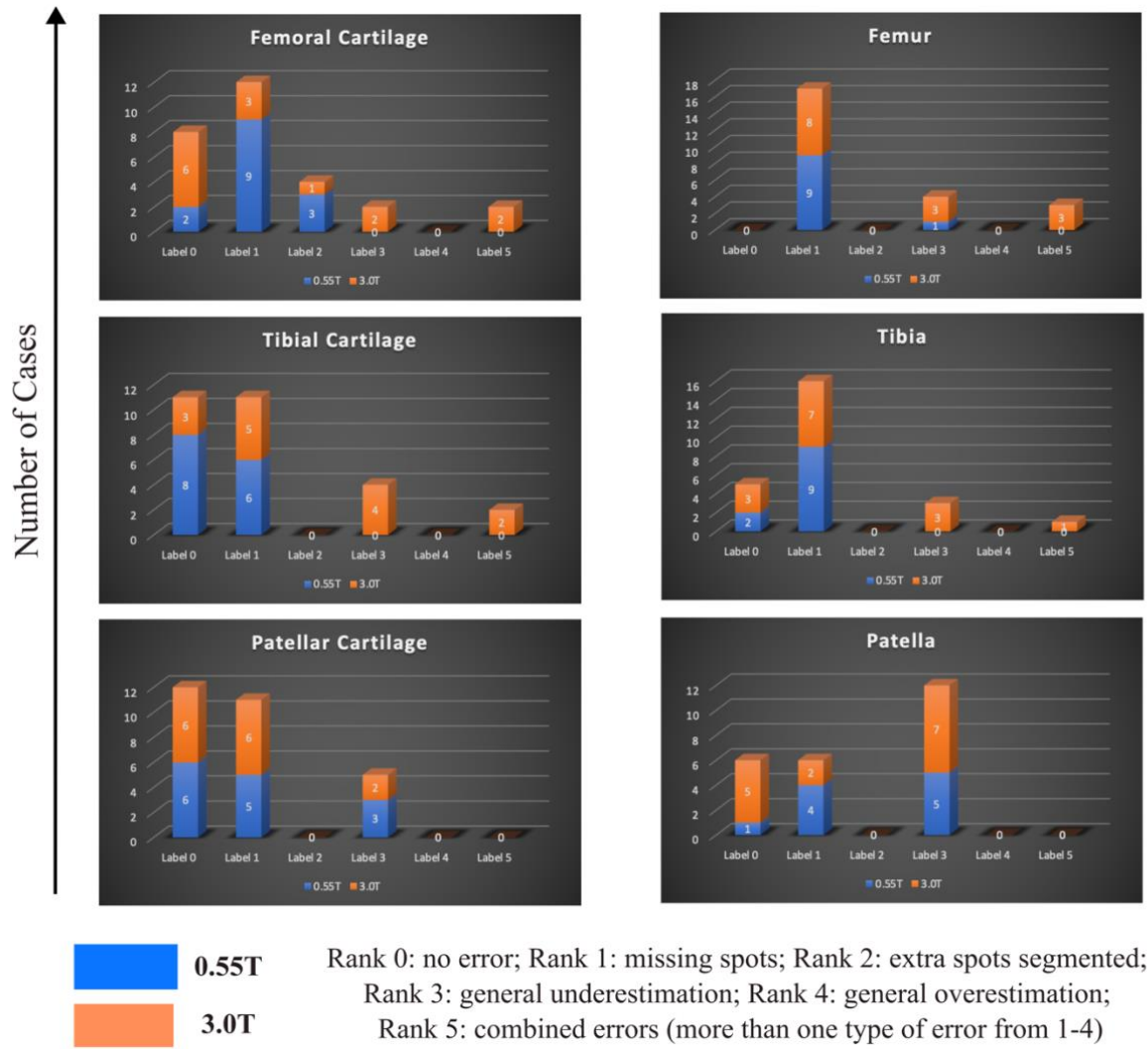

**Figure 2**: Common issues with the bone and cartilage segmentation performance at 3.0T vs. 0.55T, summarized.

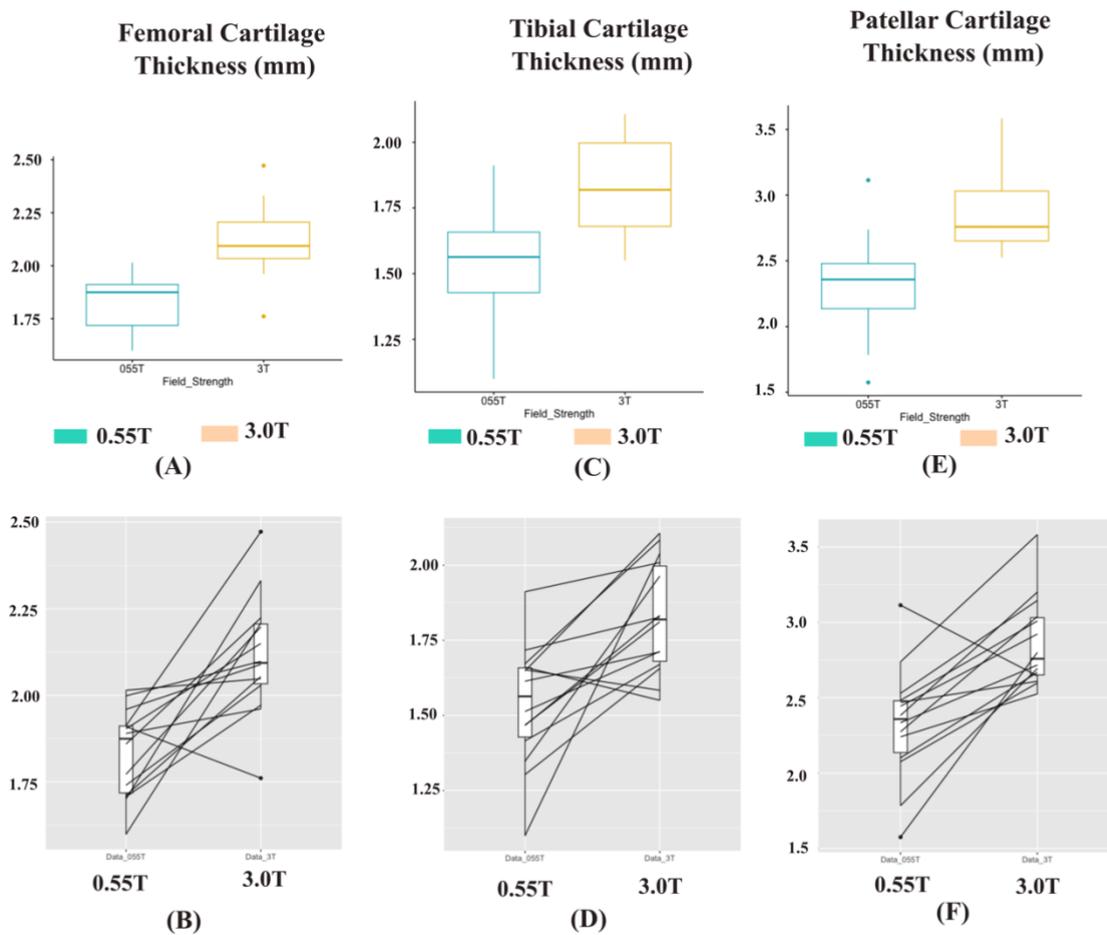

**Figure 3:** (A-B) Box plots of Femoral cartilage thickness values at 0.55T and 3.0T. (C-D) Box plots of Tibial cartilage thickness values at 0.55T and 3.0T. (E-F) Box plots of Patellar cartilage thickness values at 0.55T and 3.0T. A trend of general underestimation of cartilage thickness at 0.55T is noted from B, D, and F. The mean differences are minuscule: 0.279 (Femoral), 0.290 (Tibial), and 0.543 (Patellar) respectively.